\pdfoutput=1
\documentclass[5p,times,procedia]{elsarticle}
\usepackage{ecrc}
\volume{00}

\firstpage{1}

\journalname{Procedia CIRP}

\runauth{R. Real et al.}

\jid{trpro}

\usepackage{amssymb}
\usepackage[figuresright]{rotating}
\usepackage[bookmarks=false]{hyperref}
    \hypersetup{colorlinks,
      linkcolor=blue,
      citecolor=blue,
      urlcolor=blue}

\usepackage{xcolor}
\usepackage{soul}

\usepackage[capitalise]{cleveref}
\newcommand{\pref}[1]{(\cref{#1})}

\usepackage{booktabs}

\usepackage{subfig}

\begin{document}
\begin{frontmatter}



\dochead{31st CIRP Design Conference 2021 (CIRP Design 2021)}%

\title{Distinguishing artefacts: evaluating the saturation point of convolutional neural networks}


\author[a]{Ric Real\corref{*}} 
\author[a,b]{James Gopsill}
\author[a]{David Jones}
\author[a]{Chris Snider}
\author[a]{Ben Hicks}
%\ead{ric.real@bristol.ac.uk}

\address[a]{\href{https://dmf-lab.co.uk/}{Design Manufacturing Futures Lab, University of Bristol, UK}}
\address[b]{\href{https://cfms.org.uk/}{Centre for Modelling and Simulation, Bristol, UK}}

\aucores{* Corresponding author. {\it E-mail address:} ric.real@bristol.ac.uk}

\begin{abstract}
    Prior work has shown Convolutional Neural Networks (CNNs) trained on surrogate Computer Aided Design (CAD) models are able to detect and classify real-world artefacts from photographs.
    The applications of which support twinning of digital and physical assets in design, including rapid extraction of part geometry from model repositories, information search \& retrieval and identifying components in the field for maintenance, repair, and recording.
    The performance of CNNs in classification tasks have been shown dependent on training data set size and number of classes. 
    Where prior works have used relatively small surrogate model data sets ($<100$ models), the question remains as to the ability of a CNN to differentiate between models in increasingly large model repositories. 
    
    This paper presents a method for generating synthetic image data sets from online CAD model repositories, and further investigates the capacity of an off-the-shelf CNN architecture trained on synthetic data to classify models as class size increases. 
    1,000 CAD models were curated and processed to generate large scale surrogate data sets, featuring model coverage at steps of 10$^{\circ}$, 30$^{\circ}$, 60$^{\circ}$, and 120$^{\circ}$ degrees. 
    
    %Confusion Matrix not presented in this paper 
    %differentiation behavior observed in the confusion matrix. 
    
    The findings demonstrate the capability of computer vision algorithms to classify artefacts in model repositories of up to 200, beyond this point the CNN's performance is observed to deteriorate significantly, limiting its present ability for automated twinning of physical to digital artefacts. Although, a match is more often found in the top-5 results showing potential for information search and retrieval on large repositories of surrogate models. 
\end{abstract}

%\cortext[cor1]{Corresponding author. Tel.: +0-000-000-0000 ; fax: +0-000-000-0000.}

\end{frontmatter}

% \linenumbers


%\enlargethispage{-7mm}

%---------------------------------------------------------------------------------------------

\section{Introduction}
\label{main}

Recent trends in digital design, such as Twinning \cite{Jones2020} have underlined the value of rapid, or fully integrated, synchronisation between physical and digital domains to accelerate processes and enhance analytic capability.

The prototyping process comprises vital iteration between a multitude of physical and digital models \cite{Hansen2020, ulrich2003product, wynn2017perspectives}, in which both physical and digital states must be captured, aligned, and replicated across domains (i.e. updating of CAD models).

A key technical challenge thereby lies in creating a relationship between physical and digital states, such that each may be recognised as a counterpart of the other, whilst maintaining this alignment across multiple versions of a prototype. By automating the detection of physical models and searching/aligning against a digital counterpart, scope exists to reduce process time and cost via physical/digital transition.

Detection has previously utilised physical tagging (e.g. QR and barcodes) or direct scanning (e.g. photogrammetry), where these methods typically require modification to the physical prototype or generation of new digital models, recent works have demonstrated the ability of Convolutional Neural Networks (CNNs) to extract and learn distinguishable features of an artefact for classification. Thus, enabling rapid association with existing data, such as retrieving CAD models by taking a photograph of their real-world counterpart\cite{gopsill2020}. 

However, the performance of a CNN is dependent on quantity and quality of training data~\cite{Barbedo2018}, and the number of classes between which it must distinguish~\cite{deng2010}.
Where photographic training data is often sparse in prototyping, implementing a CNN becomes a significant challenge \cite{Peng}, thus the real-world viability of using CNNs for design twinning is not known. 

This paper investigates (a) the performance and implementation challenges associated with CNN use at increasing scale, and (b) the viability of 'off-the-shelf' CNN architectures for artefact classification. We consider an artefact to be a designed object, whose form can be distinguished and classified. 

The paper proceeds to present related works in the field of CNN use in design \pref{sec:rel}, followed by a methodology for testing CNN scalability \pref{sec:methodology}. Results are then reported \pref{sec:res} and a discussion ensues with respect to CNN's and their ability to support twinning activities in design \pref{sec:disc}. The paper concludes by detailing the key findings from the study.

%-------------------------------------------------------------------------------------------------------
%   Related work 

\section{Related work}
\label{sec:rel}

\textbf{CNNs in design: }The application of CNNs to design is a rapidly emerging field of inquiry with many potential impacts across Engineering Design.
For example, \cite{zaki2016} trained a CNN on multi-view renders of 3D geometry and used the resulting CNN to classify other 3D geometry.
A potential use case for this is the matching of similar parts across product families with a view to reduce the part variety in an organisations supply chain.
\cite{maturana2015} have sought to augment depth mapped images with Neural Networks to develop voxel-based approximations of objects within a scene, with potential application to design via providing a means to describe the locations in which products are used and deployed.
\cite{gopsill2020} is seeking to democratise design by using a CNN as an information search and retrieval tool for large model repositories, such as Thingiverse and MyMiniFactory. 
The CNN enables users to simply take a photo of the item that they wish to print and return the closest matching result in the repositories dataset.
\cite{gopsill2021} recently demonstrated CNN's being able to emulate mathematical and user perceptions on shape and form.
These could be used to check for conformation to brand identity as well as potential infringements on others.
It could also twin user feedback into the design process where the CNN will act as market feedback providing real-time assessment of designs as a designer is working on their product's design.
While such examples demonstrate the exploration of value of CNNs in design, there remain questions around their performance, and challenges in their implementation.

\textbf{Surrogate models for dataset generation: }The challenge of acquiring datasets, with thousands of images required per artefact, has previously been a limiting factor in exploring the utility of CNNs in large-scale classification tasks. With the context of design prototyping requiring sufficient data to distinguish between 100s of prototype or component versions, this presents a systemic obstacle of particular importance.

Recent work has shown the viability of synthetic image dataset generation, whereby a surrogate CAD model is processed and rendered with computer graphics software to generate two-dimensional artefact representations~\cite{Su,gopsill2020}. Image composition, for example lighting, surfaces, background, and appearance can additionally be tuned to replicate real-world environmental features ~\cite{Zhang2020,Peng,Sarkar}.

This method presents an opportunity to leverage existing virtual models for large-scale classification, whereby a CNN is trained on synthetic images from a surrogate model data set; thus mitigating the need for `real-world' photographs, and prior restrictions in data acquisition \cite{gopsill2020}. To-date however, the effect of model repository size on the efficacy of this approach has not been investigated. Where CNN detection accuracy will typically decrease as repository size increases, the extensive training sets producible via surrogate models give scope to substantially increase performance.

%-------------------------------------------------------------------------------------------------------
%    Methodology

\section{Methodology}
\label{sec:methodology}

The study followed a 5 step computational process to determine the scaling behaviour of a CNN, for the purposes of twinning physical artefacts with their digital counterpart in large repositories.

% that has been used in previous twinning studies.
%to differentiate between models and the resource requirements to achieve it.

\begin{enumerate}
    \item Dataset curation
    \item Surrogate model curation
    \item CNN selection
    \item CNN preparation
    \item Evaluation
\end{enumerate}

%
% start: David to do
%
\subsection{Dataset curation}

The data set used consisted of 1,049 STL files collected from the model sharing website MyMiniFactory.com \footnote{URL: \url{MyMiniFactory.com} Last visited: 2020-11-25, Author: MyMiniFactory}. 
MyMiniFactory is a website that allows users to upload, share and sell 3D models, and also provide an API to programmatically interact with the 3D model database.
Between June 10th and June 15th 2020, the API was queried using a Python script with the search terms `\emph{spare part}', `\emph{3d printer}', and  `\emph{accessibility}'. Additionally, a filter was applied to restrict results to only those under a Creative Commons free to re-use license.
A total of 1,514 files were downloaded, consisting predominantly of STL files (1,112 files) and a range of CAD files.
A small number of those 1,112 STL files were corrupted and removed, leaving a data set of 1,049 usable files.

\subsection{Surrogate model curation}

Where prior work has used relatively small datasets ($<100$ models), this work leverages surrogate models based on existing CAD geometry to enable CNN use in the design scenario.

The open source 3D computer graphics software Blender (version 2.8) was used to create photo-realistic renders of the artefacts.
Blender is widely used throughout the computer games and film industry to create and render 3D graphics and animation. 
It also provides a Python library, which in the case of the work presented in this paper, allowed for scripted model loading, creation of lighting, camera positions, and image rendering.

\begin{figure}[t]
    \centering
    \includegraphics[width=0.9\columnwidth]{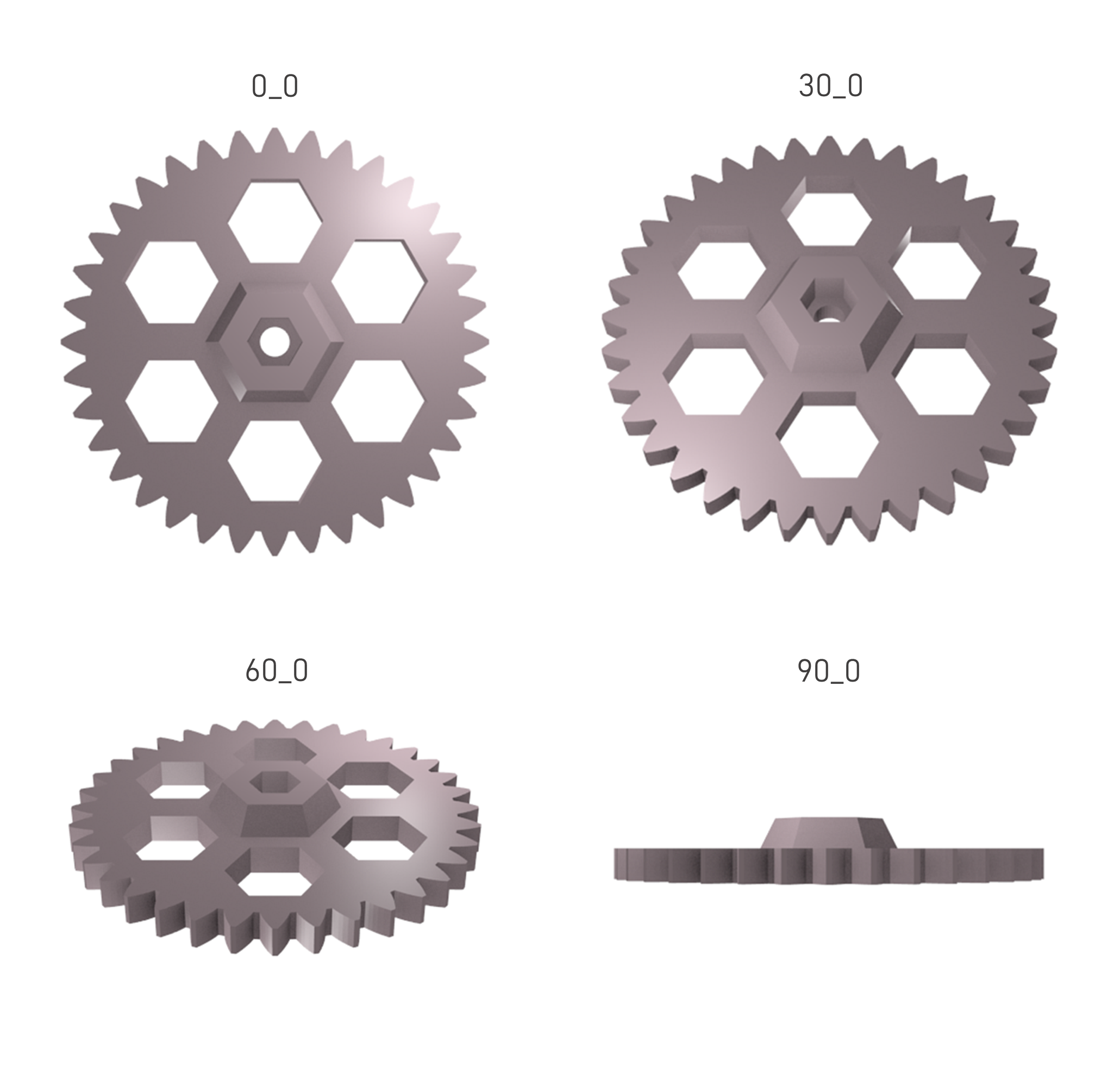}
    \caption{Sample synthetic renders at 30 Degree view representations}
    \label{fig:gears}
\end{figure}

Fixed scene lighting was used for all models with two lamps positioned (xyz) at (L1) 4.0, 4.0, 4.0 and (L2) -4.5, -4.5, -4.5, illuminating models as per \cref{fig:gears}.

Model size was also normalised and scaled via re-scaling of the bounding box, such that the artefact could be represented fully in a 540px x 540px rendered output. 

\begin{figure}[t]
    \centering
    \includegraphics[width=0.85\columnwidth]{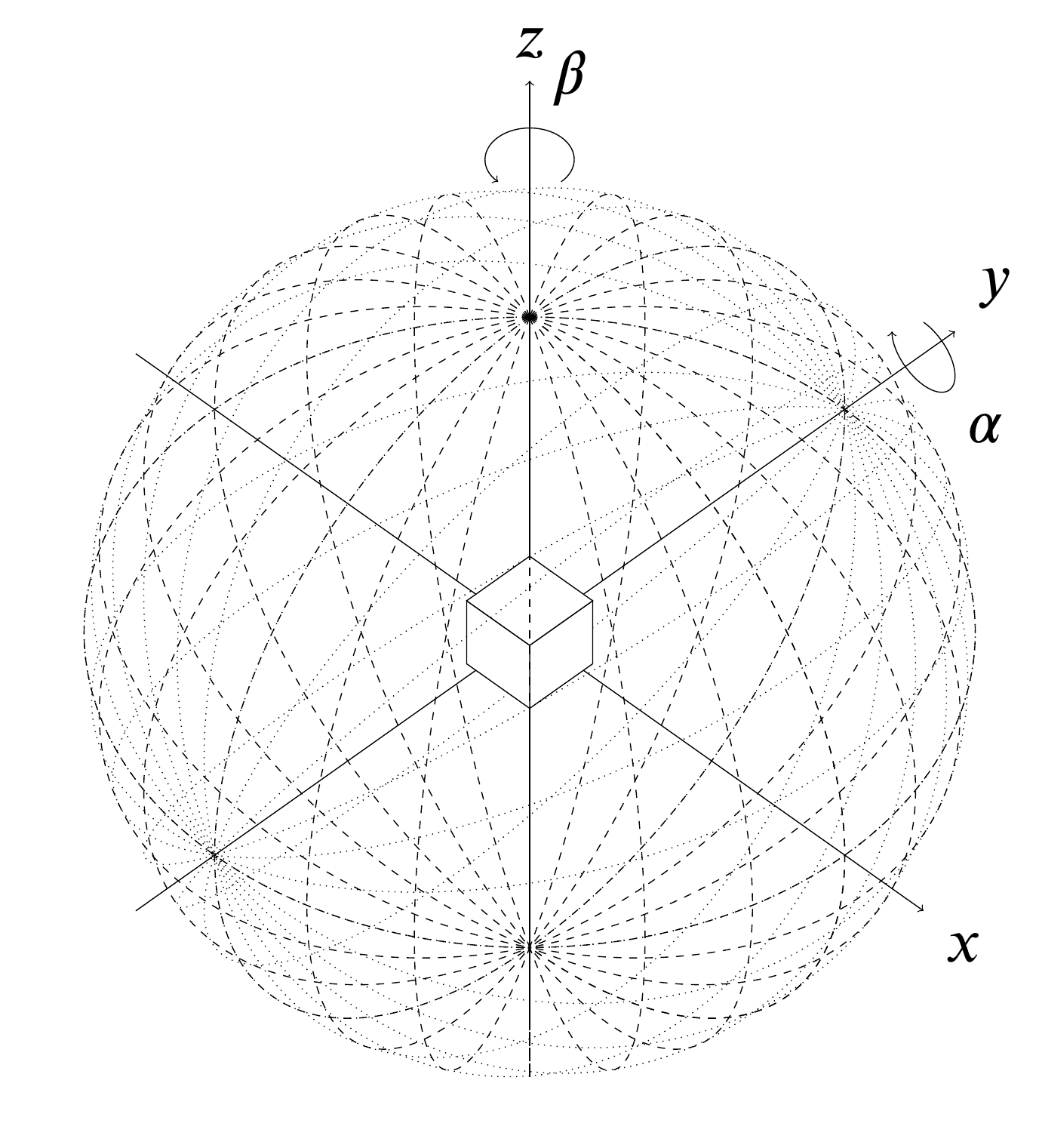}
    \caption{Camera sphere (From:~\cite{gopsill2020})}
    \label{fig:camera_sphere}
\end{figure}

Camera positions were set using longitude and latitude angles, allowing the camera to be rotated around the model (see \cref{fig:camera_sphere}).
To generate a range of images the longitude and latitude angles were set at 10$^{\circ}$, 30$^{\circ}$, 60$^{\circ}$, and 120$^{\circ}$ degrees, resulting in four data sets of 684, 84, 24 and 6 renders/images per artefact respectfully. 
A summary is provided in \cref{tab:renders}.

% Table for Rendering time of synthetic data sets from 5 to 1000 classes 
\begin{table}[t]%\vspace*{4pt}
    \caption{Rendering time per artefact (RTPA), and total dataset generation time (TDGT) for a repository size of 1000 surrogate STL models, rendered at varying degree steps (DST).} 
    \footnotesize
    \centering
    \begin{tabular}{llll}
        \toprule
        \textbf{DST}  & \textbf{Renders PA (Total)} & \textbf{RTPA ({\it{mm:ss}})} & \textbf{TDGT}({\it{dd:hh:mm}})  \\
        \midrule
        10$^{\circ}$ & 684 (684,000) & 06:00 & 04:04:00 \\ 
        30$^{\circ}$ & 84 (84,000) & 02:30 & 01:17:40 \\
        60$^{\circ}$ & 24 (24,000) & 00:12 & 00:03:20\\
        90$^{\circ}$ & 12 (12,000) & 00:06 & 00:01:40 \\
        120$^{\circ}$ & 6 (6000) & 00:03 & 00:00:50\\
        \bottomrule
    \end{tabular}
\label{tab:renders}
\end{table}

\subsection{CNN selection}

This study utilised an AlexNet `off-the-shelf' CNN architecture.
The network, designed for a 1000 category classification capacity, established prominence in computer vision research with a breakthrough Top-5 performance in the 2012 ILSVRC contest, significantly outperforming prior architectures.
Through its substantial documentation in literature and relative ease of use, AlexNet remains a widely used CNN architecture with diverse applications.  
Specifically to this context, \cite{gopsill2020} showed AlexNet to outperform other top classification architectures, GoogleNet, and ResNet, achieving 94.9\% accuracy with a six class surrogate model training data set when tested against photographs of the physical artefact.

\begin{figure*}[t]%\vspace*{4pt}
    \includegraphics[width=\textwidth]{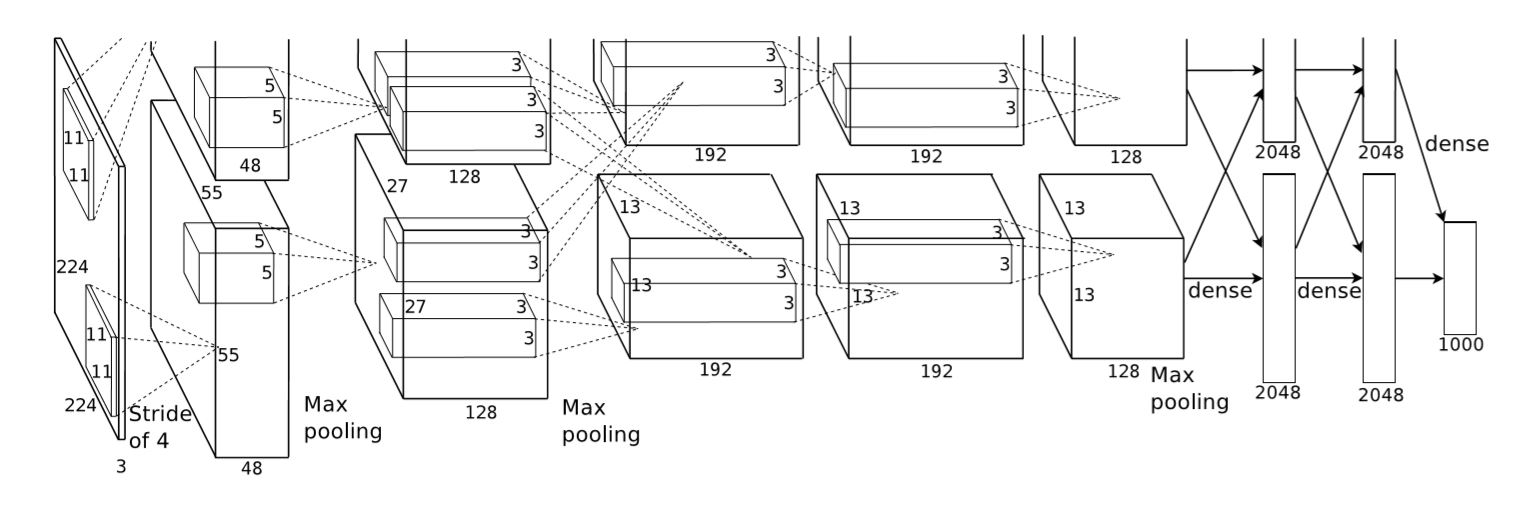}
    \caption{AlexNet network architecture, containing five convolutional and three fully-connected layers with a 1000-way SoftMax output.}
    \label{fig:alexnet}
\end{figure*}

AlexNet's architecture consists of 62 million parameters and a 1000-way output classifier~\cite{Krizhevsky}.
It comprises eight layers; the first five are convolutional and remaining three layers are fully connected \pref{fig:alexnet}. 
The final layer, a 1000-way softmax classifier, outputs a probability distribution across 1000 class labels. 
The network input is a 3 channel RGB image of dimensions 227 x 227 pixels. 
To reduce over-fitting, complexity measures including dropout and data augmentation are introduced.   
The CNN was recreated using TensorFlow 2.0 and Keras deep learning API. 
A Nvidia 8GB RTX 2060 GPU was used for training. 

\subsection{CNN preparation}

Three steps were taken to prepare the CNN following guidance in literature and early experimentation with the surrogate data.
These are:

\begin{itemize}
    \item Forming the data pipeline and applying image augmentation prior to parsing through the CNN.
    \item Configuring the Hyperparameters.
    \item Training the CNN.
\end{itemize}

\subsubsection{Data pipeline and Augmentation}

\begin{figure}[t]
    \centering
    \includegraphics[width=0.85\linewidth]{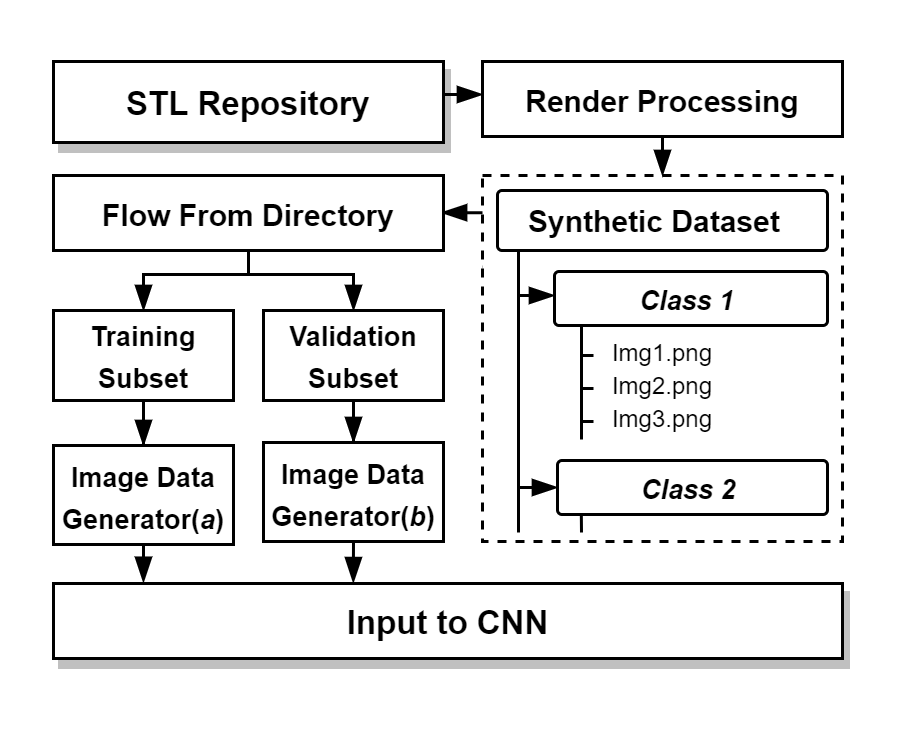}
    \caption{Synthetic Data generation pipeline, with generator \textbf{\emph{(a)}} augmenting image outputs.}
    \label{fig:pipeline}
\end{figure}

From the Keras data preprocessing utilities a pipeline was defined to generate batches of tensor image data with real-time augmentation.
The CNN was fed images from the surrogate data set using flow from directory method, automatically inferring class (artefact name) from the sub-directory structure.
30\% of images per class were partitioned to a validation subset for training.
Additionally, parallel image data generators were created for both training and validation data subsets, allowing training data to be generated with augmentation and data shuffling whilst preserving artefact representations in the validation generator~\cite{Wang}. This process is shown in figure \ref{fig:pipeline}.

\subsubsection{Hyperparamaters}

To improve training stability and CNN generalisation, a mini-batch size of 32 was used to estimate the error gradient~\cite{Masters}. Respectively, a learning rate of 1e-06 was applied to scale the value by which weights were updated in back propagation. These values were found to be effective with batch size, and the generated synthetic image data.
An epoch count of 200 was defined to observe training performance over time and allow sufficient space for model convergence.   
 
\subsubsection{Training}

CNN parameters are trained with weights and biases initialised on the active data set. `Adam' is used as the optimisation algorithm, and categorical cross entropy set as the loss function~\cite{Kingma2015}. 
This approach is elected over transfer learning with pre-trained networks to explore causality between our surrogate model data structures and the CNN's performance. 
Additionally, hyperparameters are kept constant across data sets and training iterations. 

\section{Evaluation}

CNN performance was evaluated using Keras metrics on three of five generated data sets (30$^{\circ}$, 60$^{\circ}$, 90$^{\circ}$), measuring classification accuracy and loss for training and validation steps, Top-5 categorical accuracy, and training time.
Initial test data showed performance to be inconclusive for the 120$^{\circ}$ data set, and the 10$^{\circ}$ data set to be computationally inefficient, therefore these were omitted from the study. 
Results from the candidate data sets were logged and graphed to visualise performance across dimensions of interest; accuracy, number of classes, model coverage (number of Images per class), and training time.       

%------------------------------------------------------------------------------------------------------
%    Results 

\section{Results}
\label{sec:res}

This section reports the results from our study into a CNN trained on a CAD surrogate model dataset. 
These have been reported with respect to training performance and classification accuracy.

\subsection{Training Performance}

Training time \pref{tab:training_time} shows that higher quantities of artefact in a given training set correlate to longer CNN training times. Time is observed to increase exponentially across data sets, however the degree reached by training time is limited by data set size.

A uniform transition exists from exponential to linear growth in training time for each of the surrogate data sets. The 30$^{\circ}$ set (featuring the highest number of training images) transitions to more linear time scaling earlier in the class range (200 classes), whilst this behaviour is observed later (400 classes) in the 60$^{\circ}$, and 90$^{\circ}$ data sets; suggesting the experimental set-up to reach a performance cap at between 16,000, and 4800 images.

\subsection{Classification accuracy}

In contrast to training time, classification accuracy decays almost exponentially as the number of classes per data set increases, showing a positive correlation between training image quantity and classification accuracy. 
Thus, accurate classification of physical artefacts can only be achieved with a small number of classes.

Top-5 performance (by which the \emph{true artefact} is named in the top 5 predictions) displays an affinity to the slope of Top-1 accuracy, although its gradient indicates a slower decay in classification performance and is more linear in nature.
At 1,000 classes, the CNN was still able to classify the matching model in the Top-5 predictions 75\% of the time.
This is a promising indication that CNNs could support design information search \& retrieval applications. It's worth noting for illustrative purpose that the average car features 30,000  components.

% Table for training time data from 5 to 1000 classes 
\begin{table}
\centering
\caption{Number of classes per data set and model training times}
\label{tab:training_time}
\scriptsize
\begin{tabular}{lllllll}
    \toprule
    & \multicolumn{3}{l}{\textbf{Training Time} ({\it{hh:mm}})} & \multicolumn{3}{l}{\textbf{Avg Epoch Time} ({\it{mm:ss}})} \\ \midrule
    \textbf{Number of Classes} & 30$^{\circ}$ & 60$^{\circ}$ & 90$^{\circ}$ & 30$^{\circ}$ & 60$^{\circ}$ & 90$^{\circ}$ \\ \midrule
    5 & 00:18 & 00:05 & 00:04 & 00:05  & 00:01  & 00:01     \\
    10 & 00:35 & 00:10 & 00:05 & 00:10 & 00:03 & 00:01     \\
    25 & 01:24 & 00:25 & 00:11 & 00:25 & 00:07 & 00:03     \\
    50 & 02:37 & 00:45 & 00:23 & 00:47 & 00:13 & 00:07     \\
    100 & 05:07 & 01:30 & 00:43 & 01:33 & 00:26 & 00:13     \\
    200 & 10:13 & 03:00 & 01:30 & 03:02 & 00:52 & 00:27     \\
    400 & 22:15 & 05:40 & 02:55 & 06:41 & 01:42 & 00:53     \\
    600 & 33:10 & 08:30 & 04:30 & 09:58 & 02:36 & 01:18     \\
    800 & 44:30 & 11:40 & 05:40 & 13:23 & 03:24 & 01:44     \\
    1000 & 55:07 & 14:20 & 07:20 & 16:33 & 04:12 & 02:06     \\
    \bottomrule
\end{tabular}
\end{table}

% Graph of model performance 30 Degree data set 
\begin{figure}[t]
    \centering
    \subfloat[$30^{\circ}$]{
        \includegraphics[width=\linewidth]{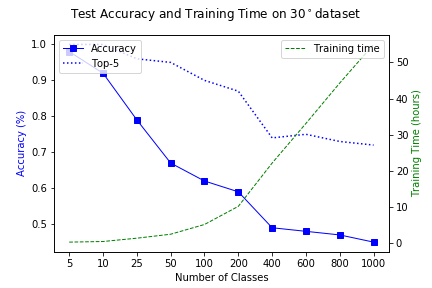}
    } \\
    \subfloat[$60^{\circ}$]{
        \includegraphics[width=\linewidth]{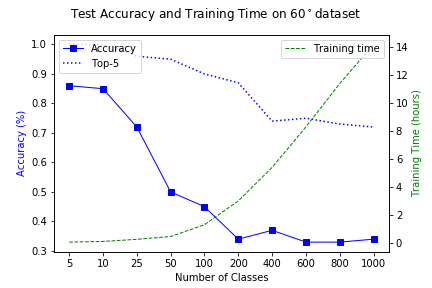}
    } \\
    \subfloat[$90^{\circ}$]{
        \includegraphics[width=\linewidth]{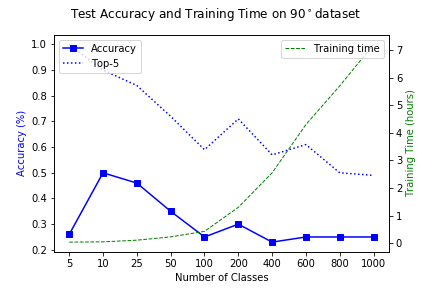}
    } \\
    \caption{Training Results}
\end{figure}

%---------------------------------------------------------------------------------------------------------
%     Discussion 

\section{Discussion and future work}
\label{sec:disc}

This study has explored the performance properties of CNNs trained on surrogate models for future application to Engineering Design.
Particularly, this work has shown the potential of adapting existing `off-the-shelf' CNN architectures to support Design activities.
This is of significance as it demonstrates that engineering organisations can use existing, general purpose AI/ML architectures rather than requiring expert consultancies to develop bespoke solutions.
%with respect to the rapid synchronisation of physical and virtual prototypes. 

Results have shown that CNNs could be applied to automated twinning between digital-physical assets if the number of artefacts to be classified remains low ($<10$). 
However, CNN performance can quickly deteriorate, this should therefore be considered in selecting suitable applications for a CNN. 

Top-5 results however show stronger performance, and suggest potential for CNN use as a rapid Search \& Retrieval tool for design information. Implementation could occur in a Search \& Retrieval tool that does not require bespoke multi-faceted search strategies as is typical in engineering due to the challenges in how one describes physical artefacts, their shape and form.
This could be significant in democratising and speeding up information search processes as design engineers are a mere photograph away from useful search results that can take them from the physical to the digital domain.

With these promising results, future work could investigate the development of bespoke CNN architectures for the activities described.
In particular, it would be interesting to apply Information Search \& Retrieval's F-score metric to the CNN and compare it to existing engineering design search strategies.

Optimisation in preparing and training the CNN is also an area to explore, ensuring development of computationally efficient and sustainable architectures that support design activities.
This would consider the number of renders required per artefact and the development of a suitable scene with appropriate lighting to accentuate artefact features as well as the application of multiple scenes.

Further work into architecture performance drop-off could yield some interesting insights into how CNNs confuse artefacts, with confusion matrices providing a means to explore this.
Is it the case that the confusion matrix of a CNN would be comparable to a similarity matrix produced by humans?
This paper has shown that there remains a wealth of research to be done to apply Machine Learning in the domain of Design, and further assess how it supports, or even hinders design processes.

\section{Conclusion}
\label{Conclusion}

Where prior work has shown the potential of CNNs to support Engineering Design activities, this paper has taken the next step in this journey by examining how CNNs scale with increasing class sizes of surrogate CAD models. 
The results have important implications in determining whether CNNs can be deployed for twinning and/or information search \& retrieval activities.

The results show that existing `off-the-shelf' CNN architectures could be re-trained and successfully deployed to twin between physical and digital domains if the number of models is low ($<10$).
The results also demonstrate the potential for CNNs to support Information Search \& Retrieval activities with the CNN being able to return a correct matching in the Top-5 for 1,000 model classes. This creates opportunity to use a single photograph to effectively retrieve virtual models of physical artefacts from large corpi.

Together, these results demonstrate the viability of CNN use in a design context, the effectiveness of the novel surrogate model training approach, and scope for future opportunities that may be realised.

\section*{Acknowledgements}

The work reported in this paper has been undertaken as part of the Twinning of digital-physical models during prototyping project. The work was conducted at the University of Bristol, Design and Manufacturing Futures Laboratory \url{(http://www.dmf-lab.co.uk)} Funded by the Engineering and Physical Sciences Research Council (EPSRC), Grant reference (EP/R032696/1).
The authors would also like to thank MiniFactory.com and their users for sharing their models.

%%-----------------------------------------------------------------------------------------------------
%%  References 
 

%The citation must be used in following style: \cite{article-minimal} \cite{article-full} \cite{article-crossref} \cite{whole-journal}.

\bibliography{paper.bib}

\begin{thebibliography}{18}
\expandafter\ifx\csname natexlab\endcsname\relax\def\natexlab#1{#1}\fi
\providecommand{\url}[1]{\texttt{#1}}
\providecommand{\href}[2]{#2}
\providecommand{\path}[1]{#1}
\providecommand{\DOIprefix}{doi:}
\providecommand{\ArXivprefix}{arXiv:}
\providecommand{\URLprefix}{URL: }
\providecommand{\Pubmedprefix}{pmid:}
\providecommand{\doi}[1]{\href{http://dx.doi.org/#1}{\path{#1}}}
\providecommand{\Pubmed}[1]{\href{pmid:#1}{\path{#1}}}
\providecommand{\bibinfo}[2]{#2}
\ifx\xfnm\relax \def\xfnm[#1]{\unskip,\space#1}\fi
%Type = Article
\bibitem[{Barbedo(2018)}]{Barbedo2018}
\bibinfo{author}{Barbedo, J.G.A.}, \bibinfo{year}{2018}.
\newblock \bibinfo{title}{{Impact of dataset size and variety on the
  effectiveness of deep learning and transfer learning for plant disease
  classification}}.
\newblock \bibinfo{journal}{Computers and Electronics in Agriculture}
  \bibinfo{volume}{153}, \bibinfo{pages}{46--53}.
\newblock \DOIprefix\doi{10.1016/j.compag.2018.08.013}.
%Type = Inproceedings
\bibitem[{Deng et~al.(2010)Deng, Berg, Li and Fei-Fei}]{deng2010}
\bibinfo{author}{Deng, J.}, \bibinfo{author}{Berg, A.C.}, \bibinfo{author}{Li,
  K.}, \bibinfo{author}{Fei-Fei, L.}, \bibinfo{year}{2010}.
\newblock \bibinfo{title}{What does classifying more than 10,000 image
  categories tell us?}, in: \bibinfo{editor}{Daniilidis, K.},
  \bibinfo{editor}{Maragos, P.}, \bibinfo{editor}{Paragios, N.} (Eds.),
  \bibinfo{booktitle}{Computer Vision -- ECCV 2010},
  \bibinfo{publisher}{Springer Berlin Heidelberg}, \bibinfo{address}{Berlin,
  Heidelberg}. pp. \bibinfo{pages}{71--84}.
%Type = Inproceedings
\bibitem[{Gopsill et~al.(2021)Gopsill, Goudswaard, Jones and
  Hicks}]{gopsill2021}
\bibinfo{author}{Gopsill, J.}, \bibinfo{author}{Goudswaard, M.},
  \bibinfo{author}{Jones, D.}, \bibinfo{author}{Hicks, B.},
  \bibinfo{year}{2021}.
\newblock \bibinfo{title}{Perceptions on shape and form}, in:
  \bibinfo{booktitle}{International Conference on Engineering Design}.
\newblock \bibinfo{note}{In Press}.
%Type = Article
\bibitem[{Gopsill and Jennings(2020)}]{gopsill2020}
\bibinfo{author}{Gopsill, J.}, \bibinfo{author}{Jennings, S.},
  \bibinfo{year}{2020}.
\newblock \bibinfo{title}{Democratising design through surrogate model
  convolutional neural networks of computer aided design repositories}.
\newblock \bibinfo{journal}{Proceedings of the Design Society: DESIGN
  Conference} \bibinfo{volume}{1}, \bibinfo{pages}{1285–1294}.
\newblock \DOIprefix\doi{10.1017/dsd.2020.93}.
%Type = Article
\bibitem[{Hansen and {\"{O}}zkil(2020)}]{Hansen2020}
\bibinfo{author}{Hansen, C.A.}, \bibinfo{author}{{\"{O}}zkil, A.G.},
  \bibinfo{year}{2020}.
\newblock \bibinfo{title}{{From Idea to Production: A Retrospective and
  Longitudinal Case Study of Prototypes and Prototyping Strategies}}.
\newblock \bibinfo{journal}{Journal of Mechanical Design}
  \bibinfo{volume}{142}, \bibinfo{pages}{031115}.
\newblock \URLprefix
  \url{https://asmedigitalcollection.asme.org/mechanicaldesign/article-pdf/doi/10.1115/1.4045385/6441489/md-19-1443.pdf},
  \DOIprefix\doi{10.1115/1.4045385}.
%Type = Article
\bibitem[{Jones et~al.(2020)Jones, Snider, Nassehi, Yon and Hicks}]{Jones2020}
\bibinfo{author}{Jones, D.}, \bibinfo{author}{Snider, C.},
  \bibinfo{author}{Nassehi, A.}, \bibinfo{author}{Yon, J.},
  \bibinfo{author}{Hicks, B.}, \bibinfo{year}{2020}.
\newblock \bibinfo{title}{{Characterising the Digital Twin: A systematic
  literature review}}.
\newblock \bibinfo{journal}{CIRP Journal of Manufacturing Science and
  Technology} \bibinfo{volume}{29}, \bibinfo{pages}{36--52}.
\newblock \DOIprefix\doi{10.1016/j.cirpj.2020.02.002}.
%Type = Inproceedings
\bibitem[{Kingma and Ba(2015)}]{Kingma2015}
\bibinfo{author}{Kingma, D.P.}, \bibinfo{author}{Ba, J.L.},
  \bibinfo{year}{2015}.
\newblock \bibinfo{title}{{Adam: A method for stochastic optimization}}, in:
  \bibinfo{booktitle}{3rd International Conference on Learning Representations,
  ICLR 2015 - Conference Track Proceedings}, \bibinfo{publisher}{International
  Conference on Learning Representations, ICLR}.
\newblock \href{http://arxiv.org/abs/1412.6980}{{\tt arXiv:1412.6980}}.
%Type = Techreport
\bibitem[{Krizhevsky et~al.()Krizhevsky, Sutskever and Hinton}]{Krizhevsky}
\bibinfo{author}{Krizhevsky, A.}, \bibinfo{author}{Sutskever, I.},
  \bibinfo{author}{Hinton, G.E.}, .
\newblock \bibinfo{title}{{ImageNet Classification with Deep Convolutional
  Neural Networks}}.
\newblock \bibinfo{type}{Technical Report}.
\newblock \URLprefix \url{http://code.google.com/p/cuda-convnet/}.
%Type = Techreport
\bibitem[{Masters and Luschi()}]{Masters}
\bibinfo{author}{Masters, D.}, \bibinfo{author}{Luschi, C.}, .
\newblock \bibinfo{title}{{REVISITING SMALL BATCH TRAINING FOR DEEP NEURAL
  NETWORKS}}.
\newblock \bibinfo{type}{Technical Report}.
\newblock \href{http://arxiv.org/abs/1804.07612v1}{{\tt arXiv:1804.07612v1}}.
%Type = Inproceedings
\bibitem[{Maturana and Scherer(2015)}]{maturana2015}
\bibinfo{author}{Maturana, D.}, \bibinfo{author}{Scherer, S.},
  \bibinfo{year}{2015}.
\newblock \bibinfo{title}{Voxnet: A 3d convolutional neural network for
  real-time object recognition}, in: \bibinfo{booktitle}{2015 IEEE/RSJ
  International Conference on Intelligent Robots and Systems (IROS)},
  \bibinfo{organization}{IEEE}. pp. \bibinfo{pages}{922--928}.
%Type = Techreport
\bibitem[{Peng et~al.()Peng, Sun, Ali and Saenko}]{Peng}
\bibinfo{author}{Peng, X.}, \bibinfo{author}{Sun, B.}, \bibinfo{author}{Ali,
  K.}, \bibinfo{author}{Saenko, K.}, .
\newblock \bibinfo{title}{{Learning Deep Object Detectors from 3D Models}}.
\newblock \bibinfo{type}{Technical Report}.
%Type = Techreport
\bibitem[{Sarkar et~al.()Sarkar, Varanasi and Stricker}]{Sarkar}
\bibinfo{author}{Sarkar, K.}, \bibinfo{author}{Varanasi, K.},
  \bibinfo{author}{Stricker, D.}, .
\newblock \bibinfo{title}{{Trained 3D models for CNN based object
  recognition}}.
\newblock \bibinfo{type}{Technical Report}.
%Type = Techreport
\bibitem[{Su et~al.()Su, Maji, Kalogerakis and Learned-Miller}]{Su}
\bibinfo{author}{Su, H.}, \bibinfo{author}{Maji, S.},
  \bibinfo{author}{Kalogerakis, E.}, \bibinfo{author}{Learned-Miller, E.}, .
\newblock \bibinfo{title}{{Multi-view Convolutional Neural Networks for 3D
  Shape Recognition}}.
\newblock \bibinfo{type}{Technical Report}.
\newblock \URLprefix \url{http://vis-www.cs.umass.edu/mvcnn.}
%Type = Book
\bibitem[{Ulrich(2003)}]{ulrich2003product}
\bibinfo{author}{Ulrich, K.T.}, \bibinfo{year}{2003}.
\newblock \bibinfo{title}{Product design and development}.
\newblock \bibinfo{publisher}{Tata McGraw-Hill Education}.
%Type = Techreport
\bibitem[{Wang and Perez()}]{Wang}
\bibinfo{author}{Wang, J.}, \bibinfo{author}{Perez, L.}, .
\newblock \bibinfo{title}{{The Effectiveness of Data Augmentation in Image
  Classification using Deep Learning}}.
\newblock \bibinfo{type}{Technical Report}.
\newblock \href{http://arxiv.org/abs/1712.04621v1}{{\tt arXiv:1712.04621v1}}.
%Type = Article
\bibitem[{Wynn and Eckert(2017)}]{wynn2017perspectives}
\bibinfo{author}{Wynn, D.C.}, \bibinfo{author}{Eckert, C.M.},
  \bibinfo{year}{2017}.
\newblock \bibinfo{title}{Perspectives on iteration in design and development}.
\newblock \bibinfo{journal}{Research in Engineering Design}
  \bibinfo{volume}{28}, \bibinfo{pages}{153--184}.
%Type = Inproceedings
\bibitem[{Zaki et~al.(2016)Zaki, Shafait and Mian}]{zaki2016}
\bibinfo{author}{Zaki, H.F.}, \bibinfo{author}{Shafait, F.},
  \bibinfo{author}{Mian, A.}, \bibinfo{year}{2016}.
\newblock \bibinfo{title}{Modeling 2d appearance evolution for 3d object
  categorization}, in: \bibinfo{booktitle}{2016 international conference on
  digital image computing: Techniques and applications (DICTA)},
  \bibinfo{organization}{IEEE}. pp. \bibinfo{pages}{1--8}.
%Type = Article
\bibitem[{Zhang et~al.(2020)Zhang, Jia and Ivrissimtzis}]{Zhang2020}
\bibinfo{author}{Zhang, X.}, \bibinfo{author}{Jia, N.},
  \bibinfo{author}{Ivrissimtzis, I.}, \bibinfo{year}{2020}.
\newblock \bibinfo{title}{{A study of the effect of the illumination model on
  the generation of synthetic training datasets}} \URLprefix
  \url{http://arxiv.org/abs/2006.08819},
  \href{http://arxiv.org/abs/2006.08819}{{\tt arXiv:2006.08819}}.

\end{thebibliography}
\bibliographystyle{elsarticle-harv}

\clearpage\onecolumn

\end{document}